\documentclass{article} 
\usepackage{iclr2024_conference,times}

\usepackage{amsmath,amsfonts,bm}

\def\eqref#1{equation~\ref{#1}}

\def\1{\bm{1}}

\DeclareMathAlphabet{\mathsfit}{\encodingdefault}{\sfdefault}{m}{sl}
\SetMathAlphabet{\mathsfit}{bold}{\encodingdefault}{\sfdefault}{bx}{n}

\newcommand{\coloredblock}[1]{\textcolor{#1}{\blacksquare}}

\usepackage{hyperref}
\usepackage{url}
\usepackage{xspace}
\usepackage{booktabs}
\usepackage{graphicx}
\usepackage{MnSymbol,bbding,pifont}
\usepackage{colortbl}
\definecolor{LightCyan}{rgb}{0.88,1,1}
\usepackage{enumitem}
\usepackage{multirow}
\usepackage{xcolor}
\definecolor{Pre-Algebra}{HTML}{F27970
}
\definecolor{Inter-Algebra}{HTML}{BB9727}
\definecolor{Algebra}{HTML}{54B345}
\definecolor{Probability}{HTML}{32B897}
\definecolor{NumTheory}{HTML}{05B9E2}
\definecolor{Calculus}{HTML}{8983BF}
\definecolor{Geometry}{HTML}{C76DA2}

\title{MAmmoTH: Building Math Generalist Models through Hybrid Instruction Tuning}

\author{
$^{\clubsuit}$Xiang Yue\thanks{Xiang Yue and Wenhu Chen are the leading authors of the paper. They contributed equally to this project.}~~, 
$^{\ddagger}$Xingwei Qu, 
$^{\dagger}$Ge Zhang, 
$^{\circ}$Yao Fu, 
$^{\mathsection}$Wenhao Huang, \\
\textbf{
$^{\clubsuit}$Huan Sun, 
$^{\clubsuit}$Yu Su, 
$^{\dagger}$Wenhu Chen$^*$
}\\
$^{\dagger}$University of Waterloo, $^{\clubsuit}$The Ohio State University, $^{\ddagger}$HKUST, $^{\circ}$University of Edinburgh, $^{\mathsection}$01.AI \\
\texttt{yue.149@osu.edu, wenhuchen@uwaterloo.ca}\\
}

\newcommand{\model}{\texttt{MAmmoTH}\xspace}
\newcommand{\modelc}{\texttt{MAmmoTH-Coder}\xspace}
\newcommand{\dataset}{\texttt{MathInstruct}\xspace}
\newcommand{\se}[1]{\textcolor{purple}{#1}}

\newcommand{\fir}[1]{\textbf{#1}}

\newcommand{\co}{$^\dagger$}

\iclrfinalcopy 
 
\begin{document}

\maketitle
\vspace{-0.7cm}
\begin{center}
    \url{https://tiger-ai-lab.github.io/MAmmoTH/}
\end{center}
\vspace{0.3cm}
\begin{abstract}
We introduce \model,  a series of open-source large language models (LLMs) specifically tailored for general math problem-solving. The \model models are trained on \dataset, our meticulously curated instruction tuning dataset. \dataset is compiled from 13 math datasets with intermediate rationales, six of which have rationales newly curated by us. It presents a unique hybrid of chain-of-thought (CoT) and program-of-thought (PoT) rationales, and also ensures extensive coverage of diverse fields in math. The hybrid of CoT and PoT not only unleashes the potential of tool use but also allows different thought processes for different math problems.  As a result, the \model series substantially outperform existing open-source models on nine mathematical reasoning datasets across all scales with an average accuracy gain between 16\% and 32\%. Remarkably, our \model-7B model reaches 33\% on MATH (a competition-level dataset), which exceeds the best open-source 7B model (WizardMath) by 23\%, and the \model-34B model achieves 44\% accuracy on MATH, even surpassing GPT-4's CoT result. Our work underscores the importance of diverse problem coverage and the use of hybrid rationales in developing superior math generalist models.
\end{abstract}

\begin{figure}[!ht]
    \centering
    \includegraphics[width=\linewidth]{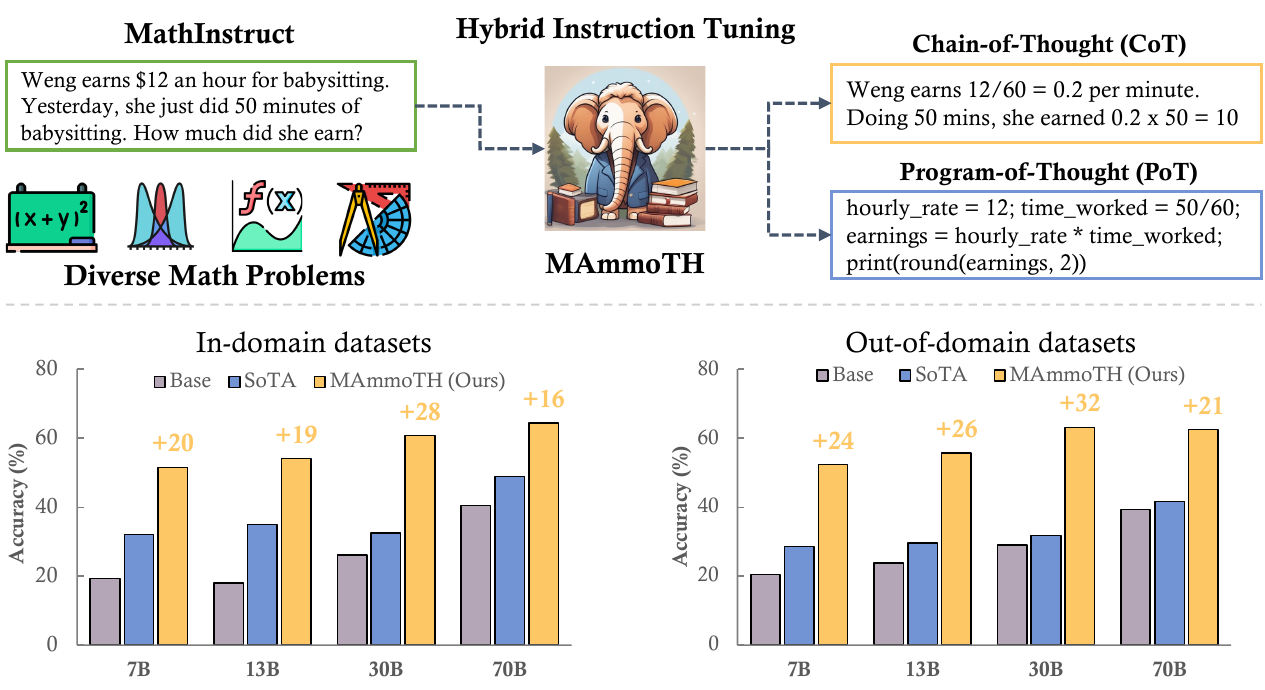}
    \caption{The superior performance of \model, a series of models instruction-tuned to solve a diverse set of mathematical problems using hybrid CoT and PoT rationales. \model significantly outperforms base and SoTA models on both in-domain and out-of-domain test sets, across all scales.}
    \label{fig:front_page_average}
\end{figure}

\section{Introduction}
This work focuses on mathematical reasoning, a critical capability of modern large language models (LLMs)~\citep{OpenAI2023GPT4TR,anil2023palm}. Despite the recent advances in this field, a noticeable gap exists between closed-source and open-source LLMs---closed-source models like GPT-4~\citep{OpenAI2023GPT4TR}, PaLM-2~\citep{anil2023palm}, and Claude 2~\citep{bai2022constitutional} dominate popular mathematical reasoning benchmarks such as GSM8K~\citep{cobbe2021training} and MATH~\citep{hendrycks2021measuring}, while open-source models like Llama~\citep{touvron2023llama1,touvron2023llama2}, Falcon~\citep{penedo2023refinedweb}, OPT~\citep{zhang2022opt} lag behind on all benchmarks by a wide margin.

Current efforts to bridge this gap are twofold: 
(1) \textit{Continued pre-training} like Galactica~\citep{taylor2022galactica} and MINERVA~\citep{lewkowycz2022solving}, which continues to train an LLM on math-related web data of more than 100B tokens. This approach improves a model's general scientific reasoning capability but incurs a high computation cost. 
(2) \textit{Dataset-specific fine-tuning} like rejection sampling fine-tuning (RFT)~\citep{yuan2023scaling} and WizardMath~\citep{luo2023wizardmath}, which fine-tunes LLMs using supervised data specific to certain datasets. Although such approaches improve in-domain performance, they cannot generalize to a wider range of math reasoning tasks beyond their fine-tuning data. For instance, both RFT and WizardMath can increase the accuracy on GSM8K~\citep{cobbe2021training} by 30\%+, one of their fine-tuning datasets, but hurt the accuracy on out-of-domain datasets like MMLU-Math~\citep{hendrycks2020measuring} or AQuA~\citep{ling2017program} by up to 10\%. 

In this paper, we aim to propose a lightweight yet generalizable math instruction-tuning approach to enhance the general (i.e., not limited to the fine-tuning tasks) mathematical reasoning capabilities of LLMs. Existing methods~\citep{luo2023wizardmath,yuan2023scaling,taylor2022galactica} primarily focus on Chain-of-Thought (CoT) approaches~\citep{wei2022chain,nye2022show} to solve math problems through step-by-step natural language descriptions. This approach excels in its generality to cover most math subjects but struggles with computation precision, and complex mathematical or algorithmic reasoning procedures (e.g., solving quadratic equation roots and calculating matrix eigenvalues).
In contrast, prompts in the format of code like Program-of-Thought (PoT) approaches~\citep{chen2022program} and PAL~\citep{madaan2022language, gao2023pal} utilize external tools (i.e., Python interpreter) to greatly simplify the math solving process. This approach advocates offloading the computation process to the external Python interpreter to solve complex mathematical and algorithmic reasoning procedures (e.g., solving quadratic equations with sympy or calculating matrix eigenvalues with numpy). However, PoT falls short in dealing with more abstract reasoning scenarios, like common-sense reasoning, formal logic, and abstract algebra, especially when there exist no built-in APIs. 

To leverage the strengths of both CoT and PoT approaches, we introduce a new math hybrid instruction-tuning dataset \dataset, which has two main characteristics: (1) \textbf{broad coverage of different math fields and complexity levels}, and (2) \textbf{hybrid CoT \& PoT rationales}. \dataset is based on seven existing math rationale datasets and six newly-curated datasets (see details in~\autoref{tab:flan}). We use \dataset to fine-tune Llama~\citep{touvron2023llama1,touvron2023llama2,roziere2023code} models of different scales ranging from 7B to 70B. The resulting \model models ( \autoref{fig:front_page_average}) demonstrate unprecedented potential in serving as math generalists. 

We evaluate \model on a spectrum of datasets, including in-domain (IND) test sets---GSM8K~\citep{cobbe2021training}, MATH~\citep{hendrycks2021measuring}, AQuA-RAT~\citep{ling2017program}, NumGLUE~\citep{mishra2022numglue}---and out-of-domain (OOD) test sets---SVAMP~\citep{patel2021nlp}, SAT~\citep{zhong2023agieval}, MMLU-Math~\citep{hendrycks2020measuring}, Mathematics~\citep{davies2021advancing}, and SimulEq~\citep{koncel2016mawps}. Compared with existing methods, our models generalize better to OOD datasets and substantially improve the performance of open-source LLMs in mathematical reasoning. Notably, on the popular competition-level MATH dataset~\citep{hendrycks2021measuring}, our 7B model can beat WizardMath (open-source MATH SoTA)~\citep{luo2023wizardmath} by 3.5x (35.2\% vs 10.7\%), and our 34B \modelc (fine-tuned on Code Llama~\citep{roziere2023code}) can even beat the result of GPT-4 (using CoT). 

We highlight our contributions from two perspectives: 
(1) From the \textbf{data engineering perspective}, we present \dataset, a high-quality math instruction tuning dataset, combining a variety of math problems and hybrid rationales. 
(2) From the \textbf{modeling perspective}, we investigate the impact of various data sources and input-output formats through training and evaluating over 50 different models and baselines ranging from 7B to 70B. Our models, including \model and \modelc, achieve substantial accuracy gains over existing open-source models.
\section{Our Approach}
\subsection{Background}
Mathematical reasoning serves as a vital gauge for assessing the ability of LLMs to execute complex multi-hop and quantitative reasoning. Previously, this has been a challenging task for neural networks, which struggle to solve even basic addition and subtraction problems~\citep{yang2023gpt}. However, recent LLMs have considerable advancements in mathematical reasoning. Key breakthroughs have been made through CoT prompting~\citep{wei2022chain,nye2022show} and PoT prompting~\citep{chen2022program,gao2023pal}. CoT prompting encourages LLMs to solve problems incrementally on a scratchpad, enhancing both accuracy and explainability in mathematical reasoning. This approach contrasts with traditional methods that generate answers directly. PoT prompting, on the other hand, formulates the intermediate reasoning process as a program, executed with an external tool like Python, to compute the answer. This method improves robustness in solving complex mathematical problems by offloading the calculations to external tools. However, most existing work~\citep{zhou2023solving} in PoT is limited to proprietary models like GPT-4~\citep{OpenAI2023GPT4TR} and Codex~\citep{chen2021evaluating}. The PoT potential of open-source models is yet to be seen. Our work aims at optimizing LLMs' CoT and PoT reasoning capabilities through instruction tuning.
\begin{table}[!t]
\centering
\small
\resizebox{\linewidth}{!}{%
\begin{tabular}{llllll}
\toprule
Training Dataset                                     & Type      & Annotation     & \# Samples        & Characteristics    & Fields    \\
\midrule
GSM8K \citep{cobbe2021training}                      & CoT       & Human          & 7K                & Grade Schol Exam   & $\coloredblock{Pre-Algebra}$  \\
{GSM8K-RFT \citep{yuan2023scaling}}                  & CoT       & Llama          & 28K              & Llama + Validated &$\coloredblock{Pre-Algebra}$  \\
{AQuA-RAT \citep{ling2017program}}                   & CoT       & Human          & 90K               & GRE/GMAT Exam      &$\coloredblock{Inter-Algebra}$       \\
{MATH \citep{hendrycks2021measuring}}                & CoT       & Human          & 7K                & Math Competition   & $\coloredblock{Pre-Algebra}$ $\coloredblock{Inter-Algebra}$ $\coloredblock{Algebra}$ $\coloredblock{Probability}$ 
$\coloredblock{NumTheory}$
$\coloredblock{Calculus}$ $\coloredblock{Geometry}$ \\
TheoremQA~\citep{chen2023theoremqa} \FiveStarConvex  & CoT       & GPT-4          & 600              & GPT4 + Validated  &$\coloredblock{Algebra}$
$\coloredblock{Probability}$
$\coloredblock{NumTheory}$
$\coloredblock{Calculus}$
$\coloredblock{Geometry}$       \\
{Camel-Math \citep{li2023camel}}                     & CoT       & GPT-4          & 50K              & GPT4 (Unvalidated)  &$\coloredblock{Algebra}$
$\coloredblock{Probability}$
$\coloredblock{NumTheory}$
$\coloredblock{Calculus}$
$\coloredblock{Geometry}$    \\
College-Math \FiveStarConvex                         & CoT       & GPT-4          & 1.8K             & GPT4 (Unvalidated)  &$\coloredblock{Algebra}$    \\
\midrule
GSM8K \FiveStarConvex                                & PoT       & GPT4           & 14K              & GPT4 + Validated    &$\coloredblock{Pre-Algebra}$      \\ 
AQuA-RAT \FiveStarConvex                             & PoT       & GPT4           & 9.7K             & GPT4 + Validated     &$\coloredblock{Inter-Algebra}$      \\
MATH \FiveStarConvex                                 & PoT       & GPT4           & 7K               & GPT4 + Validated    &$\coloredblock{Pre-Algebra}$ $\coloredblock{Inter-Algebra}$ $\coloredblock{Algebra}$ $\coloredblock{Probability}$ \\
TheoremQA \FiveStarConvex                            & PoT       & GPT4           & 700              & GPT4 + Validated    &$\coloredblock{Algebra}$
$\coloredblock{Probability}$
$\coloredblock{NumTheory}$
$\coloredblock{Calculus}$
$\coloredblock{Geometry}$      \\
{MathQA \citep{amini-etal-2019-mathqa}}              & PoT       & Human          & 25K              & AQuA-RAT Subset   &$\coloredblock{Inter-Algebra}$      \\
{NumGLUE \citep{mishra2022lila}}                     & PoT       & Human          & 13K              & Lila Annotated  &  $\coloredblock{Pre-Algebra}$        \\
\midrule
\dataset& & &260K & &$\coloredblock{Pre-Algebra}$ $\coloredblock{Inter-Algebra}$ $\coloredblock{Algebra}$ $\coloredblock{Probability}$ 
$\coloredblock{NumTheory}$
$\coloredblock{Calculus}$ $\coloredblock{Geometry}$ \\
\bottomrule
\end{tabular}
}
\caption{Overview of our \dataset. \FiveStarConvex means with NEW rationales curated by us by prompting GPT-4. We have filtered out augmented samples that have answers inconsistent with the original dataset's annotations. Different colored squares represent different fields in mathematics: $\coloredblock{Pre-Algebra}$ Pre-Algebra; 
$\coloredblock{Inter-Algebra}$ Inter-Algebra;
$\coloredblock{Algebra}$ Algebra;
$\coloredblock{Probability}$ Probability;
$\coloredblock{NumTheory}$ NumTheory;
$\coloredblock{Calculus}$ Calculus;
$\coloredblock{Geometry}$ Geometry.}
\label{tab:flan}
\end{table}

\subsection{Curating a Diverse and Hybrid Instruction Tuning Dataset}
Our study aims to compile a list of high-quality and diverse math instruction-tuning datasets, standing out with three main characteristics: (1) broad coverage of different mathematical fields and complexity levels, and (2) hybrid CoT \& PoT rationales.

\textbf{Broad Coverage of Different Math Fields and Complexity Levels:} We aim for a broad representation of math fields and complexity levels in our dataset. This ensures exposure to a diverse set of mathematical knowledge, fostering versatility in our models. Based on these criteria, we narrow down our choices to a few high-quality datasets that are widely adopted and encompass different math fields and complexity levels, such as GSM8K, MATH, AQuA, Camel, and TheoremQA. Furthermore, we notice a lack of coverage for college-level math knowledge, such as abstract algebra and formal logic, in existing datasets. To rectify this, we use GPT-4 to synthesize CoT rationales for questions in TheoremQA and create question-CoT pairs through Self-Instruct~\citep{wang2022self}, utilizing a few seed exemplars found online.

\textbf{Hybrid CoT and PoT Rationales:} Contrary to previous work~\citep{yuan2023scaling,luo2023wizardmath,lee2023platypus,wang2023far} that focus on CoT, our dataset strategically combines both. This integration enhances the dataset's versatility, catering to varying mathematical problem-solving approaches. However, most existing datasets provide limited program rationales, leading to an imbalance between CoT and PoT rationales. To fill the gap, we utilize GPT-4 to supplement the PoT rationales for selected datasets, including MATH, AQuA, GSM8K, and TheoremQA. We then filter these GPT-4 synthesized programs by comparing their executed results with human-annotated ground truth, which ensures the high quality of the added rationales.


Following these guidelines, our instruction dataset, detailed in~\autoref{tab:flan}, encompasses 260K (instruction, response) pairs, covering a wide range of core mathematical fields (arithmetic, algebra, probability, calculus, and geometry, etc.),  including hybrid CoT and PoT rationales, and offering diversity in both language and difficulty levels. This attests to its high quality and unique characteristics.

\subsection{Training Setup}
We unify all the subsets in our \dataset to conform to the structure of an Alpaca-like instruction dataset~\citep{alpaca}. This standardization ensures that the fine-tuned models can process data consistently, regardless of the original dataset formats.
We choose the open-source models Llama-2~\citep{touvron2023llama2} and Code Llama~\citep{roziere2023code} as our base models. We fine-tune these models including 7B, 13B, 34B, and 70B on \dataset, which allows us to validate our \dataset at multiple scales. We fine-tune all the models with Huggingface transformers library~\citep{wolf2019huggingface}. We use a learning rate of 2e-5 for the 7B and 13B models, and 1e-5 for the 34B and 70B models. We set the batch size at 128 and used a cosine scheduler with a 3\% warm-up period for three epochs. To efficiently train the computationally intensive 34B and 70B models, we employ DeepSpeed training with ZeRO-3 stage~\citep{rajbhandari2020zero}.

\subsection{Evaluation Setup}
\label{sec:eval_setup}
Our hybrid training enables models to solve problems using either the CoT or PoT approach. By default, the model provides the CoT solution. To switch to the PoT approach, one can add the trigger phrase ``Let's write a program to solve the problem'' following the question.

Our preliminary evaluation reveals that PoT generally outperforms CoT, notably in open-form questions like GSM8K and MATH, as programmable solutions are better at solving complex mathematical and algorithmic reasoning procedures. However, PoT struggles with abstract reasoning scenarios such as commonsense reasoning, formal logic, and abstract algebra, particularly in the absence of built-in APIs. To further combine the power of both approaches, we introduce a simple hybrid decoding strategy: The model first attempts PoT prompting. If the program is not executable, we falls back to CoT prompting. This heuristic significantly enhances our model's overall performance (see more discussions in \autoref{exp:backup_decoding}).  

\section{Experiments}

\subsection{Evaluation Datasets}
We have selected diverse evaluation datasets (Table \ref{tab:eval_datasets}), encompassing a variety of \textbf{in-domain} and \textbf{out-of-domain} samples across diverse fields of mathematics, to assess the models' capabilities in general mathematical reasoning. 

\begin{table}[!t]
\small
\centering
\resizebox{\linewidth}{!}{%
\begin{tabular}{lllll}
\toprule
Eval Dataset     & \# Samples & In-Domain? & Answer Form       & Fields                                                                                                    \\
\midrule
GSM8K \citep{cobbe2021training}           & 1319      & YES        & Open-formed         & $\coloredblock{Pre-Algebra}$                                                                          \\
MATH \citep{hendrycks2021measuring}       & 5000      & YES        & Open-formed         & $\coloredblock{Pre-Algebra}$ $\coloredblock{Inter-Algebra}$ $\coloredblock{Algebra}$ $\coloredblock{Probability}$ 
$\coloredblock{NumTheory}$
$\coloredblock{Calculus}$ $\coloredblock{Geometry}$  \\
AQuA-RAT \citep{ling2017program}          & 254       & YES        & Multi-choice & $\coloredblock{Inter-Algebra}$                                                                           \\
NumGLUE \citep{mishra2022numglue}         & 1042      & YES        & Open-formed         & $\coloredblock{Pre-Algebra}$                                                                      \\
\midrule
SVAMP \citep{patel2021nlp}                & 1000      & NO         & Open-formed         & $\coloredblock{Pre-Algebra}$                                                                                           \\
Mathematics \citep{davies2021advancing}  & 1000      & NO         & Open-formed         & $\coloredblock{Pre-Algebra}$
$\coloredblock{Inter-Algebra}$ $\coloredblock{NumTheory}$  $\coloredblock{Calculus}$                                                                    \\
SimulEq \citep{koncel2016mawps}          & 514       & NO         & Open-formed         & $\coloredblock{Inter-Algebra}$                                                                                       \\
SAT-Math \citep{zhong2023agieval}        & 220       & NO         & Multi-choice & $\coloredblock{Inter-Algebra}$ $\coloredblock{Probability}$ $\coloredblock{Geometry}$                                                               \\
MMLU-Math \citep{hendrycks2020measuring}   & 974       & NO       & Multi-choice & $\coloredblock{Algebra}$
$\coloredblock{Calculus}$ $\coloredblock{Probability}$ $\coloredblock{NumTheory}$ \\
\bottomrule
\end{tabular}
}
\caption{Comprehensive overview of our evaluation datasets, featuring a variety of in-domain and out-of-domain problems across diverse fields of mathematics. Different colored squares represent different fields in mathematics: $\coloredblock{Pre-Algebra}$ Pre-Algebra; 
$\coloredblock{Inter-Algebra}$ Inter-Algebra;
$\coloredblock{Algebra}$ Algebra;
$\coloredblock{Probability}$ Probability;
$\coloredblock{NumTheory}$ NumTheory;
$\coloredblock{Calculus}$ Calculus;
$\coloredblock{Geometry}$ Geometry.}
\label{tab:eval_datasets}
\end{table}

For the in-domain datasets, we consider GSM8K~\citep{cobbe2021training}, MATH~\citep{hendrycks2021measuring}, AQuA-RAT~\citep{ling2017program}, and NumGLUE~\citep{mishra2022numglue}. For the out-of-domain datasets, we choose SVAMP~\citep{patel2021nlp}, Mathematics~\citep{davies2021advancing}, SimulEq~\citep{koncel2016mawps}, SAT-Math~\citep{zhong2023agieval}, and MMLU-Math~\citep{hendrycks2020measuring}. The wide selection of evaluation datasets includes math problems from elementary, high school, and college levels. Some of the datasets even include formal logic and commonsense reasoning. The choice of these datasets is to ensure a comprehensive evaluation of the models' capabilities to generalize to unfamiliar situations and different math fields. The chosen evaluation datasets consist of both open-formed questions and multi-choice questions. 

\subsection{Baselines}
We partition our baselines into  the following four categories:
\begin{itemize}[leftmargin=*,itemsep=0pt, topsep=0pt, partopsep=0pt]
\item \textbf{Closed-source LLMs:} We consider 4 closed-source LLMs including GPT-4~\citep{OpenAI2023GPT4TR}, GPT-4 (Code Interpreter), PaLM-2 Unicorn~\citep{anil2023palm}, Claude-2~\citep{bai2022constitutional} and Codex~\citep{chen2021evaluating}. GPT-4, PaLM-2, and Claude-2 use CoT prompting while  GPT-4 (Code Interpreter) and Codex use PoT prompting.
\item \textbf{Llama Base:} For the base models, we consider Llama-1/2~\citep{touvron2023llama1,touvron2023llama2}, Llama-2-Chat~\citep{touvron2023llama2}.
\item \textbf{Coder Model:} To compare with different coder models, we choose Code-Llama~\citep{roziere2023code}, CodeT5+~\citep{wang2023codet5+} and CodeGen~\citep{nijkamp2023codegen}.
\item \textbf{STEM Pre-training:} We cover Galactica~\citep{taylor2022galactica} mainly to understand the performance of models specialized in STEM knowledge.
\item \textbf{Instruction Tuning:} We include Orca-Platypus~\citep{mukherjee2023orca}, Vicuna-1.5~\citep{zheng2023judging}, Tulu~\citep{wang2023far}, Platypus-2~\citep{lee2023platypus} and Guanaco~\citep{dettmers2023qlora}. We cover a wide spectrum of models trained with different types of datasets. 
\item \textbf{Dataset-Specific Tuning:} We include both RFT~\citep{yuan2023scaling} and WizardMath~\citep{luo2023wizardmath}, which specifically tune the models to adapt to GSM8K and MATH datasets. We include them to understand their generalization.
\end{itemize}
For most baselines, we choose CoT prompting to maximize their performance due to their incompetence in program generation. All the `Code Model' use PoT prompting. For GSM8K, MATH, AQuA, and NumGLUE, we will evaluate both 8-shot in-context-learning and zero-shot setups to report the highest score. For SVAMP, Mathematics, SimulEq, SAT, and MMLU, we use 5-shot in-context-learning to maintain consistency with prior work~\citep{wei2022chain,chen2023theoremqa}. Our few-shot exemplars are mostly taken from PHP\footnote{https://github.com/chuanyang-Zheng/Progressive-Hint}~\citep{zheng2023progressive}.
For \model and \modelc, we always evaluate under 0-shot setting. 
For all models, we allow a maximum sequence length of 2048 tokens for decoding. For multiple-choice questions, if the generated answer lacks an option, we map it by re-prompting the model: ``Please find the closest option to [generated answer]. The options are [options]''.

\begin{table}[!ht]
\centering
\small
\resizebox{\linewidth}{!}{%
\begin{tabular}{lllcccc|c}
\toprule
Model             & Base          & Math-SFT?    & GSM8K           & MATH            & AQuA           & NumGLUE     & Avg \\
\midrule
\multicolumn{8}{c}{Closed-source Model} \\
\midrule
GPT-4             & -             & Unknown            & 92.0\co       & 42.5\co         & 72.6\co        &  -          &  -  \\
GPT-4 (Code-Interpreter)  & -             & Unknown            & 97.0\co       & 69.7\co         & -              & -           &  -  \\
PaLM-2            & -             & Unknown            & 80.7\co       & 34.3\co         & 64.1           &  -          &  -  \\
Claude-2          & -             & Unknown            & 85.2\co       & 32.5\co         & 60.9           & -           &  -  \\
Codex (PoT)       & -             & No                 & 71.6\co       & 36.8\co        & 54.1\co        & -           &  -  \\
ART (InstructGPT) & -             & Unknown            & 71.0          & -              & 54.2           & -           &  -  \\
\midrule
\multicolumn{8}{c}{7B Parameter Model}  \\
\midrule
Llama-1           & -             & No            & 10.7\co       & 2.9\co         & 22.6           & 24.7        & 15.5  \\
Llama-2           & -             & No            & 14.6\co       & 2.5\co         & 30.3           & 29.9        & 19.3  \\
Galactica-6.7B    & GAL           & GAL-Instruct  & 10.2          & 2.2            & 25.6           & 25.8        & 15.9   \\
Code-Llama (PoT)  & -             & No            & 25.2          & \se{13.0}           & 24.0           & 26.8        & 22.2   \\
AQuA-SFT          & Llama-2       & AQuA          & 11.2          & 3.6            & \se{35.6}      & 12.2        & 15.6  \\
Llama-1 RFT       & Llama-1       & GSM8K         & 46.5\co       & 5.2            & 18.8           & 21.1        & 22.9  \\
WizardMath        & Llama-2       & GSM8K+MATH    & \se{54.9}\co  & 10.7\co        & 26.3           & \se{36.1}   & \se{32.0}  \\
\midrule
\model            & Llama-2       & \dataset      & 53.6          & 31.5           & 44.5           & 61.2        & 47.7   \\
\modelc           & Code-Llama    & \dataset      & \se{59.4}     & \se{33.4}      & \se{47.2}      & \se{66.4}   & \se{51.6}   \\
\rowcolor{LightCyan}
$\Delta$          &               &               & +5            & +21            & +12             & +30         & \fir{+20}    \\
\midrule
\multicolumn{8}{c}{13-15B Parameter Model} \\
\midrule
Llama-1           & -             & No            & 17.8\co       & 3.9\co         & 26.0           & 24.8        & 18.1   \\
Llama-2           & -             & No            & 28.7\co       & 3.9\co         & 25.1           & 8.8         & 16.6   \\
Code-Llama (PoT)  & -             & No            & 36.1          & \se{16.4}      & 28.7           & 29.2        & 27.6   \\
CodeT5+ (PoT)     & -             & No            & 12.5          & 2.4            & 20.5           & 19.4        & 13.7  \\
CodeGen+ (PoT)     & -            & No            & 12.7          & 3.4            & 24.5           & 22.5        & 15.7  \\
Vicuna-1.5        & Llama-2       & No            & 28.4\co       & 5.8            & 24.8           & 36.9        & 23.9  \\
Llama-1 RFT       & Llama-1       & GSM8K         & 52.1\co       & 5.1            & 16.1           & 24.5        & 24.4  \\
Orca-Platypus     & Llama-2       & Platypus      & 38.4          & 3.0            & 18.9           & 35.3        & 23.9  \\
Platypus          & Llama-2       & Platypus      & 25.7          & 2.5            & \se{33.4}      & \se{42.3}   & 25.9  \\
WizardMath        & Llama-2       & GSM8K+MATH    & \se{63.9}\co  & 14.0\co        & 21.2           & 40.8        & \se{34.9} \\
\midrule
\model            & Llama-2       & \dataset      & 62.0          & 34.2           & \se{51.6}           & \se{68.7}   & \se{54.1} \\
\modelc           & Code-Llama    & \dataset      & \se{64.7}     & \se{36.3}      & 46.9      & 66.8        & 53.7 \\
\rowcolor{LightCyan}
$\Delta$          &               &               & +1            & +20            & +18            & +26         & \fir{+19} \\
\midrule
\multicolumn{7}{c}{30-34B Parameter Model}  \\
\midrule
Llama-1            & -             & No            & 35.6\co       & 7.1\co        & \se{33.4}      &  28.4       & 26.1 \\
Code-Llama (PoT)   & -             & No            & 44.0          & \se{23.1}     & 25.2           &  29.3       & 30.4  \\
Llama-1 RFT        & Llama-1       & GSM8K         & \se{56.5}\co  & 7.4\co        & 18.5           &  24.3       & 26.6 \\
Galactica-30B      & GAL           & GAL-Instruct  & 41.7          & 12.7          & 28.7           &  34.7       & 29.4 \\
Platypus           & Llama-1       & Platypus      & 37.8          & 9.3           & 27.9           &  40.5       & 28.8    \\
Tulu               & Llama-2       & Tulu          & 51.0          & 10.8          & 25.5           &  \se{43.4}  & \se{32.6} \\
\midrule
\modelc            & Code-Llama    & \dataset      & \se{72.7}     & \se{43.6}     & \se{54.7}      &  \se{71.6}  & \se{60.7}  \\
\rowcolor{LightCyan}
$\Delta$           &               &               &   +16          &  +21          &  +21           & +28         & \fir{+28}   \\
\midrule
\multicolumn{7}{c}{65-70B Parameter Model}  \\
\midrule
Llama-1            & -             & No            & 50.9\co        & 10.6\co       & 35.0          &  50.2       & 36.6 \\
Llama-2            & -             & No            & 56.8\co        & 13.5\co       & 40.9          &  50.4       & 40.4 \\
Llama-2-Chat       & Llama-2       & No            & 54.9           & 18.6          & 37.0          &  51.6       & 40.5 \\
Guanaco            & Llama-2       & No            & 59.2           & 4.1           & 45.2          &  53.5       & 40.5 \\
WizardMath         & Llama-2       & GSM8K+MATH    & \se{81.6}\co   & \se{22.7}\co  & 20.0          &  48.9       & 43.3 \\
Platypus           & Llama-2       & Platypus      & 70.6           & 15.6          & \se{51.2}     &  \se{55.4}  & \se{48.1}  \\
\midrule
\model             & Llama-2       & \dataset      & \se{76.9}      & \se{41.8}     & \se{65.0}     &  \se{74.4}  & \se{64.5} \\
\rowcolor{LightCyan}
$\Delta$           &               &               & -5             & +19           & +14           & +19          & \fir{+16}  \\
\bottomrule
\end{tabular}
}
\caption{The table compiles all the \textbf{in-domain} evaluation results. Results marked as $\dagger$ are copied from other papers, which can be found on \href{https://paperswithcode.com/}{paperswithcode} leaderboards. Math-SFT? means whether the model has been instruction-tuned on any math reasoning datasets. Pink numbers highlight the highest number within the corresponding scale and dataset. Note that there does not exist a 30B+ version for Llama-2 or a 70B version for Code-Llama. }
\label{tab:ind_result}
\end{table}
\begin{table}[!t]
\centering
\small
\begin{tabular}{lccccc|c}
\toprule
Model                  & SVAMP      & Mathematics  & SimulEq   & SAT-Math   & MMLU-Math  & Avg \\
\midrule
\multicolumn{7}{c}{Closed-source Model}  \\
\midrule
GPT-4                  & 97.0\co    &  -        & -         &  95\co     &  -         &  -\\
Codex (PoT)            & 85.2\co    & -         & -         &  68\co     &  -         & - \\
\midrule
\multicolumn{7}{c}{7B Parameter Model}   \\
\midrule
Llama-1                & 24.5      & 6.2       & 4.6        &  22.7      & 30.6        &  17.7 \\
Llama-2                & 34.5      & 6.0       & 5.0        &  26.8      & 29.8        &  20.4  \\
Code-Llama (PoT)       & \se{49.4} & \se{21.7} & 3.5        &  \se{28.6} & 26.9        &  26.0   \\
Llama-1 RFT            & 21.1      & 5.1       & 11.0       &  12.5      & 21.7        &  14.3  \\
Galactica-6.7B         & 25.6      & 4.6       & 4.2        &  17.5      & 28.0        &  16.0 \\
WizardMath             & 36.1      & 9.3       & \se{12.8}  &  25.4      & \se{31.1}   &  \se{28.6}  \\
Toolformer             & 29.4\co   & -         & -          & -          & -           & - \\
\midrule
\model                 & 67.7      & 46.3      & 41.2       &  \se{42.7}      & 42.6        &  48.1    \\
\modelc                & \se{71.4} & \se{55.4} & \se{45.9}  &  40.5 & \se{48.3}   &  \se{52.3}     \\
\rowcolor{LightCyan}
$\Delta$               & +22       & +34       & +33        & +14        & +17          & \fir{+24}  \\
\midrule
\multicolumn{7}{c}{13B Parameter Model}   \\
\midrule
Llama-1                 &  34.7     &   6.9     &  5.4      &  27.7        & 30.7       & 21.0 \\
Llama-2                 &  35.1     &   11.5    &  5.8      &  32.7        & 34.4       & 23.9  \\
Code-Llama (PoT)        & \se{60.0} & \se{21.3} &  3.8      &  25.9        & 27.7       & 27.7  \\
Vicuna-1.5              &  55.7     &   10      & 6.6       &  34.0        & 34.1       & 28.1 \\
Llama-1 RFT             &  46.5     &   6.7     & 10.1      &  13.2        & 21.6       & 19.6 \\
WizardMath              &  51.9     &  14.1     & \se{14.9} &  24.5        & 32.1       & 27.5 \\
Platypus                &  55.4     &  11.4     & 7.4       &  \se{36.8}   & 35.5       & 29.3 \\
Orca-Platypus           &  56.8     &   12.6    & 7.9       &  29.5        & \se{41.6}  & \se{29.7} \\
\midrule
\model                  &  72.4     & 49.2      & 43.2      &  46.8   & 47.6  & 51.8 \\
\modelc                 & \se{73.7} & \se{61.5} & \se{47.1} &  \se{48.6}        & \se{48.3}       & \se{55.8}     \\
\rowcolor{LightCyan}
$\Delta$                & +14       & +40       & +33       &  +12          & +7         & \fir{+26}  \\
\midrule
\multicolumn{7}{c}{30-34B Parameter Model}  \\
\midrule
Llama-1                 &  48.8     & 12.8      &  11.2     & 33.4         & 39.0       & 29.0 \\
Code-Llama (PoT)        & \se{69.1} & \se{34.5} &  6.8      & 26.8         & 21.6       & \se{31.7}  \\
Llama-1 RFT             &  55.4     & 7.6       &  12.8     & 20.4         & 37.9       & 26.8  \\
Galactica-30B           &  41.6     & 11.8      &  13.2     & 37.7         & 37.9       & 28.4 \\
Tulu                    &  59.0     & 10.7      &  10.3     & 31.3         & 39.8       & 30.2 \\
Platypus                &  51.7     & 13.8      & \se{13.6} & \se{38.6}    & \se{41.0}  & \se{31.7} \\
\midrule
\modelc                 & \se{84.3} & \se{65.4} & \se{51.8} & \se{60.9}    & \se{53.8}  & \se{63.2}   \\
\rowcolor{LightCyan}
$\Delta$                &  +15         & +31          &  +38         &  +22            &  +13          & \fir{+32} \\
\midrule
\multicolumn{7}{c}{65-70B Parameter Model}  \\
\midrule
Llama-1                 & 55.3      & 14.2      & 15.2       & 37.4        & 44.1       & 33.2\\
Llama-2                 & 63.8      & 20.5      & 14.0       & 51.3        & 47.1       & 39.3 \\
Llama-2-Chat            & 71.5      & 19.2      & 21.7       & 44.1        & 46.9       & 40.6\\
WizardMath              & \se{71.8} & 17.1      & \se{37.9}  & 13.2        & 27.4       & 33.4 \\
Guanaco                 & 66.8      & 17.8      & 20.2       & 50.0        & 47.3       & 40.4 \\
Platypus                & 51.8      & \se{26.3} & 21.7       & \se{55.9}   & \se{52.5}  & \se{41.6} \\
\midrule
\model                  & \se{82.4} & \se{55.6} & \se{51.4}  & \se{66.4}   & \se{56.7}  & \se{62.5} \\
\rowcolor{LightCyan}
$\Delta$                & +11       & +29       & +14         & +11        & +4         & \fir{+21}  \\
\bottomrule
\end{tabular}
\caption{The table compiles all the \textbf{out-of-domain} evaluation results.  Results marked as $\dagger$ are copied from other papers, which can be found on \href{https://paperswithcode.com/}{paperswithcode} leaderboards.}
\vspace{-10pt}
\label{tab:ood_results}
\end{table}

\subsection{Main Results}
We report our in-domain and out-of-domain results in~\autoref{tab:ind_result} and~\autoref{tab:ood_results} respectively. Overall, we can see that \model and \modelc are able to outperform the SoTA model at different scales. In general, the performance gain for OOD datasets is more significant than IND datasets. These results show us the potential of our models as a mathematical generalist. On several datasets, \modelc-34B and \model-70B are even surpassing closed-source LLMs. 

From~\autoref{tab:ind_result}, we can observe that our main competitors for IND datasets are WizardMath~\citep{luo2023wizardmath} and Platypus~\citep{lee2023platypus}. WizardMath's training is heavily rooted in GSM8K and MATH datasets. Therefore, WizardMath's results are highly competitive on these two datasets. However, the dataset-specific training can be detrimental to OOD datasets like AQuA. In contrast, Platypus fine-tunes LLMs on a wide range of text and math reasoning datasets. it improves the open-source SoTA on several datasets. Similarly, \model can achieve universal improvement across the board. A major observation is that \model is particularly strong at solving more complex math problems in MATH, where the gain of our model over WizardMath (open-source SoTA on MATH) can exceed 25\% at different scales. 

From~\autoref{tab:ood_results}, we can observe that our main competitor for OOD datasets is Platypus~\citep{lee2023platypus}. Similar to in-domain results, Platypus is able to yield gains over the baseline models universally across the board, especially on the MMLU-Math dataset, which is tied with \model-70B. It is worth noting that the performance gains of our model on OOD datasets are even more significant than on in-domain datasets. This demonstrates our models' remarkable generalizability to unseen math problems. Notably, \model-7B also boosts the CoT performance of WizardMath-7B greatly on MMLU-Math by 9\%, which contains a substantial number of questions beyond the subjects we covered in our training dataset.
\begin{figure*}[!t]
    \centering
\includegraphics[width=0.95\linewidth]{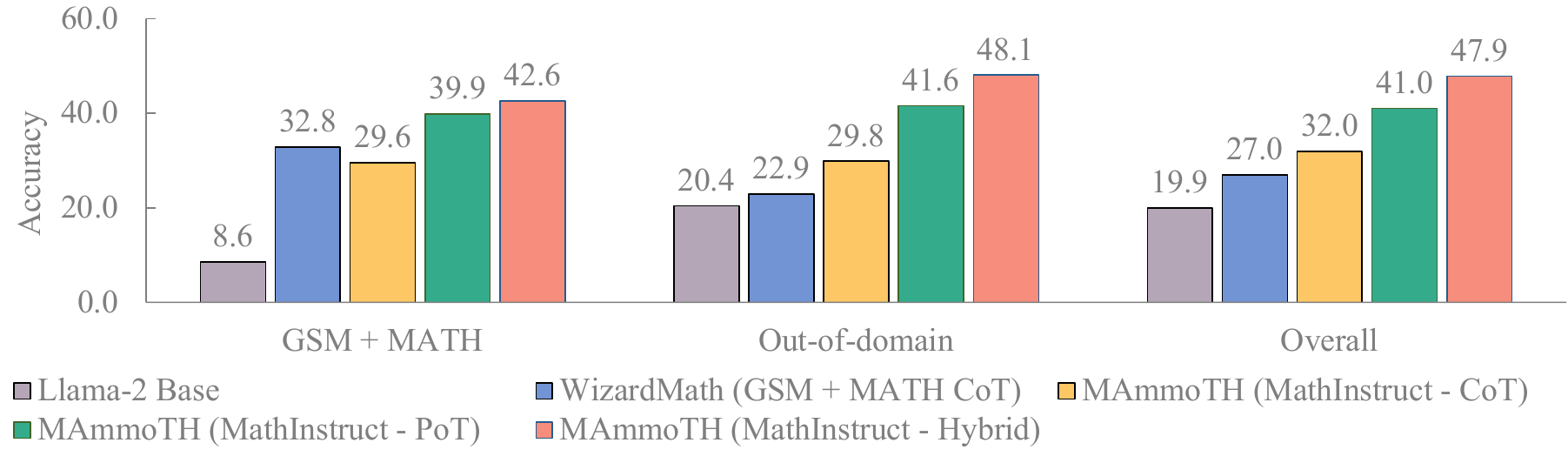}
    \caption{Investigation of the influence of CoT \& PoT hybrid training on the 7B Llama-2 model. ``Out-of-domain'' refers to the five datasets detailed in \autoref{tab:eval_datasets}. Key insights include: 1) The SoTA model, utilizing dataset-specific CoT fine-tuning on GSM and MATH, displays strong performance within its domains but struggles in OOD scenarios; 2) Diverse data
    sources in \dataset enable better math generalist model; 3) Fine-tuning on the PoT subsets generally outperforms fine-tuning on the CoT subsets; 4) Hybrid training yields the best-performing model. The breakdown results on each dataset can be found in Appendix \autoref{tab:ablation_prompt}.  }
    \label{fig:ablation_prompt}
    \vspace{-10pt}
\end{figure*}

\textbf{Comparison between Different Base Models.}
In our experiments, we experimented with both Llama-2 and Code-Llama as the base models. From the two tables, we can observe that Code-Llama is consistently better than Llama-2, especially on OOD datasets. The gap between \model and \modelc can even reach up to 5\%. Surprisingly, the average performance on OOD datasets of \modelc (34B) is actually higher than \model (70B). We believe \modelc benefits greatly from the continuous code training of Code-Llama, which not only enhances the PoT capabilities but also improves Llama's general reasoning skills. 

\subsection{Ablation Study on Data Source}
\textbf{Ablation of the Data Source.}
In order to better understand what factors contribute to the great gain of \model over existing baselines, we set up a group of control experiments in~\autoref{fig:ablation_prompt}. We study the following setups:  

(1) \model (\dataset - CoT): This experiment aims to understand how much our curated CoT data could improve the generalization over the SoTA model WizardMath~\citep{luo2023wizardmath}  trained specifically on GSM + MATH. As can be seen, while sacrificing accuracy on GSM + MATH by 3\%, our CoT subset fine-tuning improves the overall nine-dataset accuracy from 27\% to 32\%. 

(2) \model (\dataset - PoT): This experiment aims to understand the advantage of our PoT subset. As can be observed, our PoT subset fine-tuning can significantly improve the overall accuracy from 27\% to 41\%. This ablation reflects the importance of unlocking the program generation capabilities of our model.

(3) \model (\dataset - Hybrid): We further combine CoT and PoT as the hybrid training data to achieve the best overall performance of 47.9\%. This combined gain comes from two aspects: 
\begin{itemize}[leftmargin=*,itemsep=0pt, topsep=0pt, partopsep=0pt]
    \item The CoT subset helps maintain generic language-based reasoning skills to handle scenarios where PoT cannot handle well, e.g., abstract reasoning multi-choice questions in AQuA and MMLU.
    \item The PoT subset can teach the model how to utilize Python APIs to solve complex math problems with high precision, e.g., the MATH problems requiring complex computation.
\end{itemize}
We put some case studies in \autoref{sec:case_study} to demonstrate the respective advantages of PoT and CoT in solving different types of math problems. To summarize, we attribute our substantial gain to: 1) diverse data sources covering different math fields and complexity levels and 2) a hybrid of CoT \& PoT instruction tuning and decoding strategy.

\paragraph{Influence of Major Subsets.}
Given the diverse sources of \dataset used in training \model, it is important to understand how each dataset contributes to the overall performance of the model. We focus on five significant subsets: GSM8K, MATH, Camel, AQuA and NumGLUE. We conduct an experiment gradually adding each dataset into training and compare the performance with the one fine-tuned on the whole \dataset. As we can see from \autoref{tab:ablation_dataset}, when the data is not very diverse in training at the beginning (e.g., GSM8K only), the overall generalization performance is very bad: the model only fits in-distribution data and struggles to answer questions beyond GSM questions. And when gradually adding other major subsets, besides seeing the improvements on its own test sets overall, we could observe \model becomes a better math generalist. 

These results underscore the significant impact of diverse data sources on \model performance, a core aspect of making \model a math generalist.
The results also provide valuable insights for future data curation and collection efforts (e.g., we should always collect diverse data and avoid collecting only specific types of data).

To help understand the contribution of the 6 newly curated datasets as shown in \autoref{tab:flan}, we remove them from \dataset, and train a model on the existing data. As shown in the last two rows of \autoref{tab:ablation_dataset}, our new curated data substantially improves the performance on many datasets and leads to a 9\% overall increase, which reflects the importance of the NEWLY curated dataset.

\label{exp:backup_decoding}
\paragraph{Influence of Hybrid Decoding.}
To demonstrate the effectiveness of the hybrid decoding method, we conduct an experiment as outlined in \autoref{sec:eval_setup}. By default, we initially attempt the PoT decoding method for a given question. If it fails to generate an executable query, we then transition to the CoT decoding method. The performance of different decoding methods (CoT, PoT, and Hybrid) is shown in \autoref{tab:backup_decoding}. This hybrid decoding improves performance on every test set, showcasing that our model can effectively leverage the strengths of both CoT and PoT decoding strategies.

\begin{table}[!t]
\small
\centering
\resizebox{\linewidth}{!}{%
\begin{tabular}{@{}lccccccccc|c}
\toprule
Training Data                   & GSM & MATH   & AQuA & NumG    & SVA   & Mat         & Sim     & SAT  & MMLU   & \textbf{AVG}   \\ \midrule
-                               & 14.6  & 2.5  & 30.3 & 29.9    & 34.5  & 6.0         & 5.0     & 26.8 & 29.8   & \textbf{-25.3} \\
\midrule
G & 56.6 & 9.2 & 24.4 & 32.1 & 65.4 & 20.5 & 12.3 & 27.2 & 25.2 & \textbf{-22.7} \\
G + M & 58.1 & 28.2 & 26.0 & 34.7 & 64.8 & 50.1 & 17.1 & 28.6 & 28.4 & \textbf{-19.5} \\
G + M + C & 57.4 & 28.5 & 26.2 & 37.5 & 65.3 & 50.4 & 17.7 & 29.3 & 28.7 & \textbf{-19.2} \\
G + M + C + A & 57.5 & 29.1 & 46.9 & 42.2 & 65.8 & 49.6 & 32.7 & 42.3 & 43.1 & \textbf{-4.8} \\
G + M + C + A + N & 56.5 & 28.9 & 38.2 & 63.7 & 64.1 & 47.9 & 40.8 & 38.6 & 44.5 & \textbf{-3.4} \\\midrule
Existing Data                   & 31.4  & 18.4 & 40.3 & 53.3    & 61.8  & 27.9        & 45.6    & 32.7 & 38.4   & \textbf{-9.0}  \\ 
\midrule
\dataset & 53.6 & 31.5 & 44.5 & 61.2 & 67.7 & 46.3 & 41.2 & 42.7 & 42.6 & \textbf{47.9} \\ \bottomrule
\end{tabular}
}
\caption{Influence of different major subsets in \dataset based on Llama-2 7B. G: GSM8K, M: MATH, C: Camel, A: AQuA, N: NumGLUE. ``Existing data'': the subset of \dataset in \autoref{tab:flan} by excluding all the NEW rationales curated by us. We shorten Mathematics as Mat, SimulEq as Sim, NumGLUE as NumG, and SVAMP as SVA to save space. }
\label{tab:ablation_dataset}
\end{table}

\section{Conclusion}
In this paper, we propose a novel math instruction tuning approach to activate open-source LLMs' mathematical reasoning capabilities. Through a comprehensive study, we show that our models can outperform the SoTA performance at different scales by a huge margin. Our models benefit massively from: 1) the broad coverage of different math fields and complexity levels, and 2) a hybrid of CoT and PoT training. Our instruction tuning dataset contains 260K samples, which makes fine-tuning highly affordable even for academic labs. Our work paves the road for future studies to activate LLMs' core capabilities in specialized domains. 


\clearpage
\appendix
\section{Related Work}
\subsection{Mathematical Reasoning Datasets}
Our work builds upon the existing mathematical reasoning literature. Early on, mathematical reasoning is mostly focused on solving synthetic basic math problems like AddSub~\citep{hosseini2014learning} and other arithmetic reasoning datasets~\citep{koncel2015parsing,roy2015solving,patel2021nlp}. Later on, more difficult math word problem datasets~\citep{cobbe2021training,amini-etal-2019-mathqa,ling2017program,hendrycks2021measuring} have been proposed to focus on addressing realistic math word problems. NumGLUE~\citep{mishra2022numglue} and LiLA~\citep{mishra2022lila} compile the existing literature to build a more diversified dataset collection. However, these datasets are mostly focused on grade school math problems. To further test LLMs' limits in addressing more complex math problems, MMLU~\citep{hendrycks2020measuring} includes college math problems in its evaluation suite. More recently, ~\citep{chen2023theoremqa,wang2023scibench} have proposed to tackle more challenging college-level science and math problems. Our instruction tuning dataset is built upon existing work to include a diversified collection of math problems from different subfields. 

\subsection{Reasoning with Large Language Models}
LLMs have demonstrated great capabilities to reason with the help of Chain-of-Thought prompting~\citep{wei2022chain,kojima2022large,wang2022selficlr}. ~\citet{suzgun2022challenging} have shown that CoT can already surpass human performance on challenging BIG-Bench tasks. Later on, several other works~\citep{drozdov2022compositional,zhou2022least,nye2022show, wang2022iteratively,wang-etal-2023-towards,li2023making,wang2023making,yu2023metamath} also propose different approaches to utilize LLMs to solve reasoning tasks by allowing intermediate steps. ReAct~\cite{yao2022react} proposes to leverage external tools like search engines to enhance LLM reasoning skills.  Another trend is to enable LLMs' capabilities to use programs as thought processes like PoT~\citep{chen2022program}. Some follow-up works include self-critic~\citep{gou2023critic}, self-eval~\citep{xie2023decomposition}, plan-and-solve~\citep{wang2023plan}. These methods propose to enhance LLMs' capabilities to solve math problems with PoT. Self-critic~\citep{gou2023critic} and self-eval~\citep{xie2021explanation} both adopt self-evaluation to enhance the robustness of the generated program. Plan-and-solve~\citep{wang2023plan} instead adopts more detailed planning instructions to help LLMs create a high-level reasoning plan. These methods all prove to bring decent improvements over PoT.

\subsection{Instruction Tuning in Language Models}
Instruction tuning is part of a line of work designed to “align” language models with more useful objectives and human preferences. The instruction tuning step is seen as a major step to activate LLMs' certain capabilities to respond to human instructions. Previously, instruction tuning is mainly focused on enhancing LLMs' general-purpose instruction following abilities. Since 2021, CrossFit~\citep{ye2021crossfit} and NaturalInstruction~\citep{wang2022super}, FLAN~\citep{wei2021finetuned} and T0~\citep{sanh2021multitask} are amongst the first wave of instruction tuning effort to understand LLMs' generalization capabilities. Later on, FLAN-v2~\citep{chung2022scaling,longpre2023flan} have been proposed to understand the effect of scaling up the instruction datasets to understand its impact on model performance. These approaches mainly adopt human-annotated datasets to build the instruction following dataset. More recently, multiple works~\citep{wang2022self,xu2023wizardlm,peng2023instruction,zhou2023lima,wang2023far} propose to utilize synthetic instruction following data distilled from GPT-3/4 to align open-source LLMs. The most similar effort to ours is Platypus~\citep{lee2023platypus} which aims to utilize a domain-specialized dataset to construct a small-scale instruction following dataset to enhance LLMs' reasoning capabilities.

\newpage
\section{Case Study}
\label{sec:case_study}
\begin{figure}[!ht]
    \centering
    \includegraphics[width=1.0\linewidth]{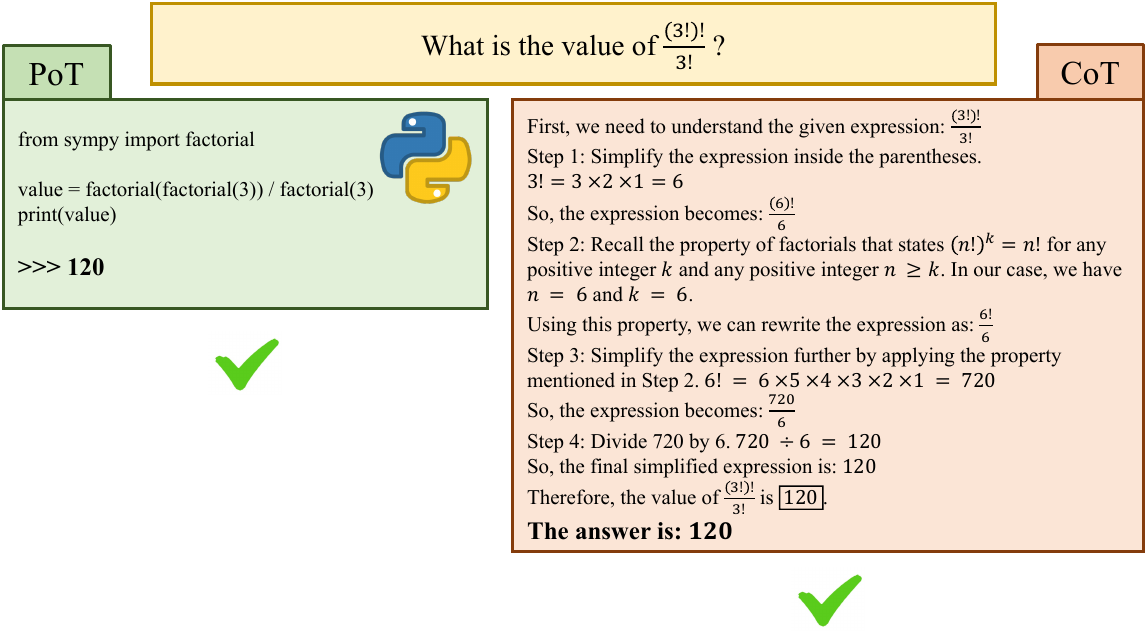}
    \caption{Example 1: PoT and CoT can both solve the problem, however, CoT gives a very tedious solution to derive the answer.}
    \label{fig:example1}
\end{figure}

\begin{figure}[!ht]
    \centering
    \includegraphics[width=1.0\linewidth]{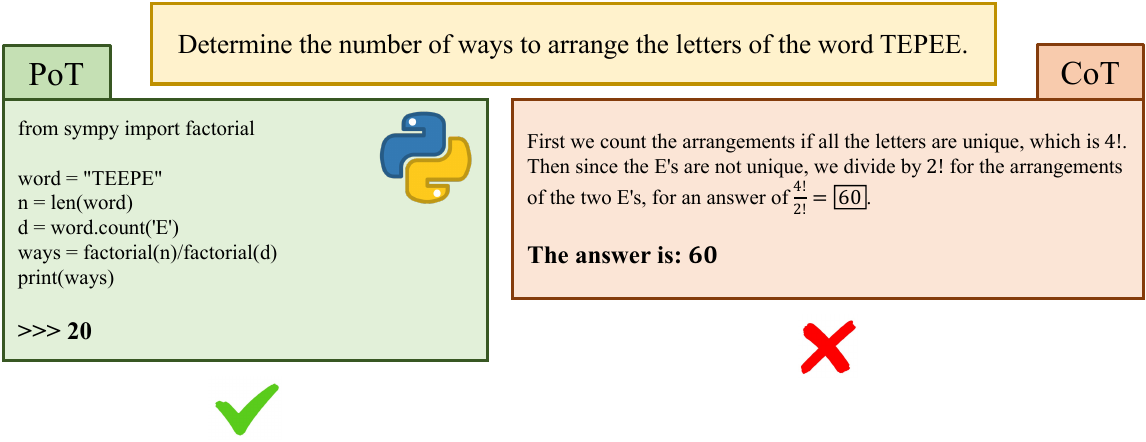}
    \caption{Example 2:  PoT generates the correct solution while CoT fails.}
    \label{fig:example2}
\end{figure}

\begin{figure}[!ht]
    \centering
    \includegraphics[width=0.9\linewidth]{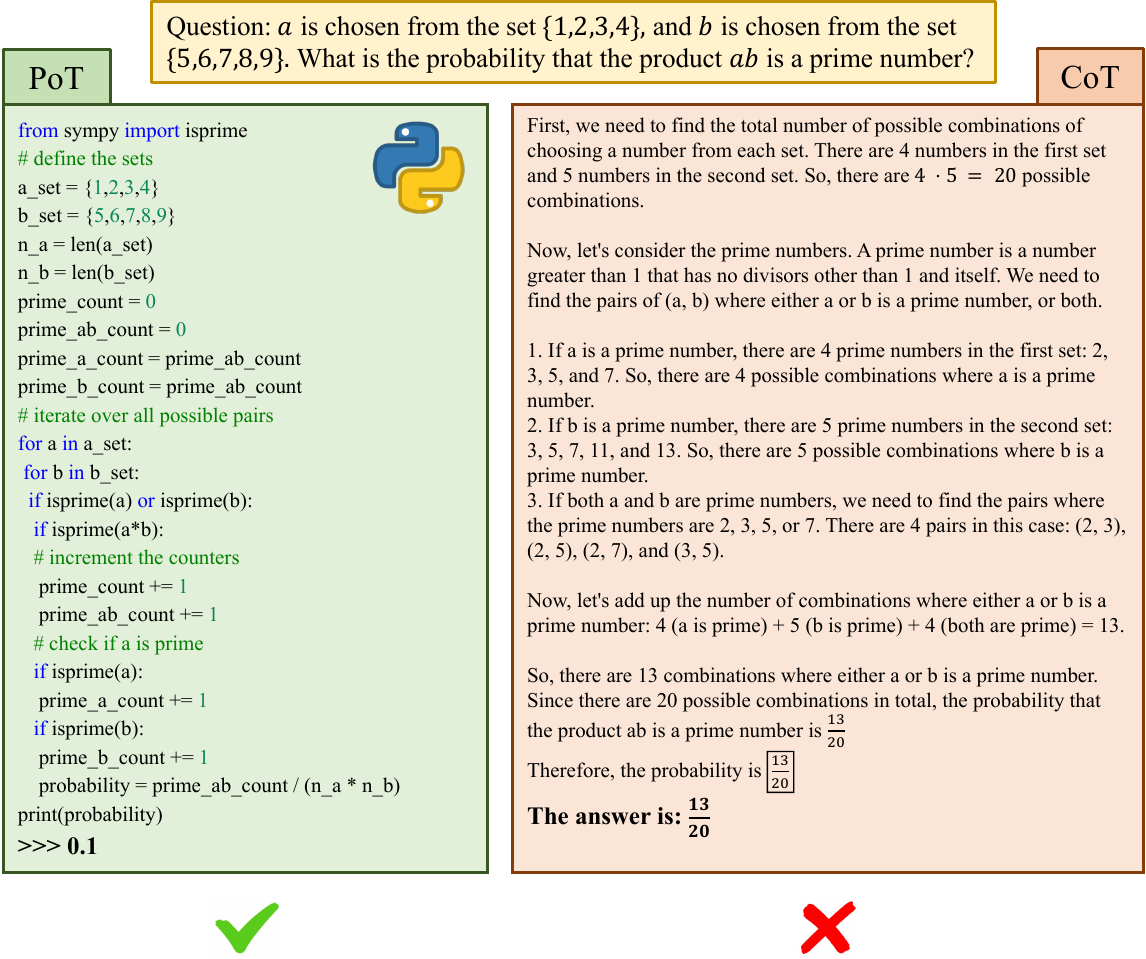}
    \caption{Example 3: PoT generates the correct solution while CoT fails.}
    \label{fig:example3}
\end{figure}

\begin{figure}[!ht]
    \centering
    \includegraphics[width=0.9\linewidth]{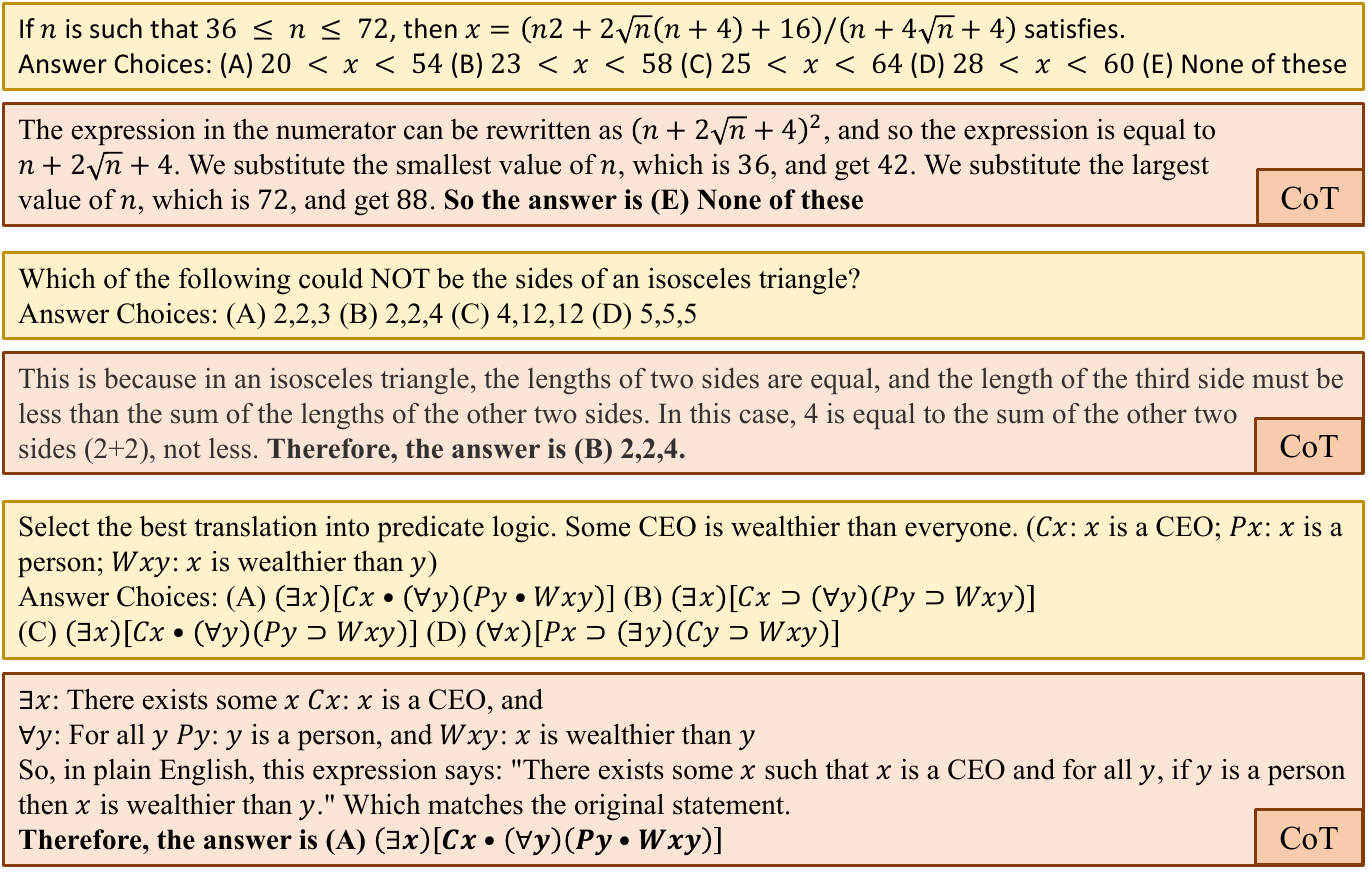}
    \caption{Example 4: Some types of questions (e.g., formal logic) are hard to be solved by PoT but could be handled by CoT.}
    \label{fig:example4}
\end{figure}

We conduct a comparison between our PoT results vs. CoT results in~\autoref{fig:example1}, ~\autoref{fig:example2} and ~\autoref{fig:example3}. In the first example, even though PoT and CoT can both solve the problem, CoT gives a very tedious solution to derive the answer. Such solution is not only slow but also unstable. In the second and third case, we can further see the advantages of PoT over CoT by utilizing external tools and Python packages to greatly simplify the solution. \autoref{fig:example4} shows some types of questions (especially the formal logic question) that are not easily handled by programs. In order to address these types of questions, CoT is a better choice.

\newpage
\section{Limitations}
Despite their training on a diverse set of mathematical rationale datasets, the \model models might exhibit limitations when faced with problems outside their primary domain of expertise like mathematical analysis, complex analysis, graph theory, numerical analysis, etc. Thus, our models are not suitable for solving more complex problems in these fields. Also, they have not been trained with proof-type problems, thus their theorem-proving capability is also limited. In the future, we would like to expand the models' skill set to cover more fields and theorem-proving problems.

There is also a risk of the \model models generating potentially harmful, offensive, or biased content, especially if they are asked to answer questions beyond math. The \model series could be misused for malicious purposes, such as spreading misinformation or probing sensitive topics. Developers should conduct safety testing and tuning tailored to their specific applications before deploying any \model model. While we have made every effort to ensure the cleanliness and purity of our training data, we cannot guarantee absolute perfection. It is unlikely but not impossible that some inappropriate questions slipped through the curation process. 

Future work may continue to explore how to further improve the robustness and generalizability of \model in mathematical reasoning. For example, recent work identifies ``sycophancy'' and ``Clever Hans effect'' in reasoning: LLMs cannot maintain truthful solutions to reasoning tasks when challenged by the user's absurdly invalid arguments and critiques~\citep{wang2023can}. Potential methods to improve the models' reasoning robustness could involve the exploration of synthetic data intervention methods as explored in ~\citep {wei2023simple}.

\begin{table}[!t]
\resizebox{\linewidth}{!}{%
\begin{tabular}{@{}lcccccccccc@{}}
\toprule
 Model & GSM & MATH & AQuA & NumG & SVA & Mat & Sim & SAT & MMLU & Overall \\ \midrule
Base & 14.6 & 2.5 & 30.3 & 29.9 & 34.5 & 6.0 & 5.0 & 26.8 & 29.8 & 19.9 \\ \midrule
WizzardMath & 54.9 & 10.7 & 26.3 & 36.1 & 36.1 & 9.3 & 12.8 & 25.4 & 31.1 & 27.0 \\ \midrule
\model (\dataset - CoT) & 49.2 & 9.9 & 42.2 & 37.1 & 48.5 & 9.5 & 17.3 & 34.1 & 39.8 & 32.0 \\
\model  (\dataset - PoT) & 50.8 & 28.9 & 28.6 & 52.7 & 65.0 & 46.7 & 42.0 & 25.9 & 28.3 & 41.0 \\
\model  (\dataset) & 53.6 & 31.5 & 44.5 & 61.2 & 67.7 & 46.3 & 41.2 & 42.7 & 42.6 & 47.9 \\ \bottomrule
\end{tabular}%
}
\caption{Breakdown results of Figure \ref{fig:ablation_prompt}. Investigation of the influence of CoT \& PoT hybrid training on the 7B Llama-2 model.  }
\label{tab:ablation_prompt}
\end{table}

\begin{table}[!t]
\resizebox{\linewidth}{!}{%
\begin{tabular}{@{}llcccccccccc@{}}
\toprule
Model & Decoding & GSM & MATH & AQuA & NumG & SVA & Mat & Sim & SAT & MMLU & AVG \\ \midrule
\multirow{3}{*}{\model-7B} & CoT & 50.5 & 10.4 & 43.7 & 44.0 & 47.3 & 9.2 & 18.9 & 32.7 & 39.9 & 33.0 \\
 & PoT & 51.6 & 28.7 & 43.3 & 52.3 & 65.1 & 41.9 & 48.2 & 39.1 & 44.6 & 46.1 \\
 & \textbf{Hybrid} & \textbf{53.6} & \textbf{31.5} & \textbf{44.5} & \textbf{61.2} & \textbf{67.7} & \textbf{46.3} & \textbf{41.2} & \textbf{42.7} & \textbf{42.6} & \textbf{47.9} \\ \midrule
\multirow{3}{*}{\modelc-7B} & CoT & 22.4 & 7.9 & 36.2 & 36.0 & 37.0 & 8.2 & 7.2 & 32.7 & 34.6 & 24.7 \\
 & PoT & 58.8 & 32.1 & 47.2 & 57.1 & 71.1 & 53.9 & 44.6 & 40.0 & 47.8 & 50.3 \\
 & \textbf{Hybrid} & \textbf{59.4} & \textbf{33.4} & \textbf{47.2} & \textbf{66.4} & \textbf{71.4} & \textbf{55.4} & \textbf{45.9} & \textbf{40.5} & \textbf{48.3} & \textbf{52.0} \\ \midrule
\multirow{3}{*}{\model-13B} & CoT & 56.3 & 12.9 & 45.3 & 45.6 & 53.8 & 11.7 & 22.4 & 43.6 & 42.3 & 37.1 \\
 & PoT & 61.3 & 32.6 & 48.8 & 59.6 & 72.2 & 48.5 & 40.3 & 46.8 & 45.4 & 50.6 \\
 & \textbf{Hybrid} & \textbf{62.0} & \textbf{34.2} & \textbf{51.6} & \textbf{68.7} & \textbf{72.4} & \textbf{49.2} & \textbf{43.2} & \textbf{46.8} & \textbf{47.6} & \textbf{52.9} \\ \midrule
\multirow{3}{*}{\modelc-13B} & CoT & 32.1 & 10.2 & 40.6 & 36.2 & 43.0 & 9.6 & 10.1 & 40.9 & 36.6 & 28.8 \\
 & PoT & 64.3 & 35.2 & 46.8 & 54.2 & 73.2 & 60.0 & 44.2 & 48.2 & 48.2 & 52.7 \\
 & \textbf{Hybrid} & \textbf{64.7} & \textbf{36.3} & \textbf{46.9} & \textbf{66.8} & \textbf{73.7} & \textbf{61.5} & \textbf{47.1} & \textbf{48.6} & \textbf{48.3} & \textbf{54.9} \\ \midrule
\multirow{3}{*}{\modelc-33B} & CoT & 34.3 & 11.6 & 39.0 & 36.2 & 44.6 & 10.8 & 10.9 & 46.4 & 42.9 & 30.7 \\
 & PoT & 72.3 & 42.8 & 53.8 & 59.6 & 84.0 & 64.7 & 50.6 & 58.6 & 52.7 & 59.9 \\
 & \textbf{Hybrid} & \textbf{72.7} & \textbf{43.6} & \textbf{54.7} & \textbf{71.6} & \textbf{84.3} & \textbf{65.4} & \textbf{51.8} & \textbf{60.9} & \textbf{53.8} & \textbf{62.1} \\ \midrule
\multirow{3}{*}{\model-70B} & CoT & 72.4 & 21.1 & 57.9 & 58.9 & 71.6 & 20.0 & 31.9 & 57.3 & 52.1 & 49.2 \\
 & PoT & 76.7 & 40.1 & 60.2 & 64.3 & 81.7 & 55.3 & 45.3 & 64.1 & 53.5 & 60.1 \\
 & \textbf{Hybrid} & \textbf{76.9} & \textbf{41.8} & \textbf{65.0} & \textbf{74.4} & \textbf{82.4} & \textbf{55.6} & \textbf{51.4} & \textbf{66.4} & \textbf{56.7} & \textbf{63.4} \\ \bottomrule
\end{tabular}%
}
\caption{Influence of different decoding methods on each dataset.}
\label{tab:backup_decoding}
\end{table}

\end{document}